\documentclass[runningheads]{llncs}
\usepackage{graphicx}
\usepackage{times}
\usepackage{latexsym}
\usepackage{multirow}
\usepackage{makecell}

\usepackage{amsmath}
\usepackage{amssymb}
\usepackage{url}
\usepackage{booktabs} 
\usepackage{graphicx, multirow}
\usepackage{CJKutf8}
\usepackage{subfigure}

\begin{document}
\begin{CJK*}{UTF8}{gbsn}
\title{Neural Chinese Word Segmentation with Dictionary Knowledge}

%
%
\author{Junxin Liu\inst{1} \and
Fangzhao Wu\inst{2} \and
Chuhan Wu\inst{1} \and
Yongfeng Huang*\inst{1} \and
Xing Xie\inst{2}}
\authorrunning{Junxin et al.}
%
\institute{Department of Electronic Engineering, Tsinghua University, Beijing, China\\
\email{\{ljx16,wuch15\}@mails.tsinghua.edu.cn}\\
\email{yfhuang@mail.tsinghua.edu.cn*}\and
Microsoft Resesrch Asia, Beijing, China\\
\email{\{fangzwu,Xing.Xie\}@microsoft.com}}

\maketitle              
\begin{abstract}

Chinese word segmentation (CWS) is an important task for Chinese NLP.
Recently, many neural network based methods have been proposed for CWS.
However, these methods require a large number of labeled sentences for model training, and usually cannot utilize the useful information in Chinese dictionary.
In this paper, we propose two methods to exploit the dictionary information for CWS.
The first one is based on pseudo labeled data generation, and the second one is based on multi-task learning.
The experimental results on two benchmark datasets validate that our approach can effectively improve the performance of Chinese word segmentation, especially when training data is insufficient.

\keywords{Chinese word segmentation  \and Dictionary \and Neural network.}
\end{abstract}

\section{Introduction}

Different from English texts, in Chinese texts there is no explicit delimiters such as whitespace to separate words.
Thus, Chinese word segmentation (CWS) is an important task for Chinese natural language processing~\cite{chen2015long,zhang2018neural}, and an essential step for many downstream tasks such as POS tagging~\cite{zheng2013deep}, named entity recognition~\cite{luo2016empirical}, dependency parsing~\cite{chen2014feature,zhang2013chinese} and so on.

Since a Chinese sentence is usually a sequence of Chinese characters, Chinese word segmentation is usually modeled as a sequence labeling problem~\cite{xue2003chinese,zhang2018neural}.
Many sequence modeling methods such as hidden Markov model (HMM)~\cite{eddy1996hidden} and conditional random field (CRF)~\cite{lafferty2001conditional} have been applied to the CWS task.
A core problem in these sequence modeling based CWS methods is building the feature vector for each character in sentences.
In traditional CWS methods these character features are constructed via manual feature engineering~\cite{peng2004chinese,zhao2006effective}.
These handcrafted features need a large amount of domain knowledge to design, and the size of these features is usually very large~\cite{chen2015long}.

In recent years, many neural network based methods have been proposed for CWS~\cite{chen2015long,zhang2016transition,zhang2018neural,zheng2013deep}.
For example, Peng et al.~\cite{peng2017multi} proposed to use Long Short-Term Memory Neural Network (LSTM) to learn the character representations for CWS and use CRF to jointly decode the labels.
However, these neural network based methods usually rely on a large number of labeled sentences.
For words which are scarce or absent in training data, these methods are very difficult to correctly segment the sentences that contain these words~\cite{zhang2018neural}.
Since these words are in large quantity, it is very expensive and even unpractical to improve the coverage of these words via annotating more sentences.
Luckily, many of these words are well defined in existing Chinese dictionaries.
Thus, Chinese dictionaries have the potential to improve the performance of neural network based CWS methods and reduce the dependence on labeled data~\cite{zhang2018neural}.


In this paper we propose to incorporate the dictionary information into neural network based CWS approach in an end-to-end manner without any feature engineering.
More specifically, we propose two methods to incorporate the dictionary information for CWS.
The first one is based on pseudo labeled data generation, where we build pseudo labeled sentences by randomly sampling words from Chinese dictionaries.
The second one is based on multi-task learning.
In this method we introduce another task named Chinese word classification (i.e., classifying a sequence of Chinese characters based on whether they can form a Chinese word), and jointly train this task with CWS by sharing the parameters of neural networks.
We conducted extensive experiments on two benchmark datasets.
The experimental results validate that our methods can effectively improve the performance of CWS, especially when training data is insufficient.

\section{Related Work}

In recent years, many neural network based methods have been proposed for Chinese word segmentation~\cite{chen2015long,zhang2016transition,zhang2018neural,zheng2013deep}.
Most of these methods model CWS as a sequence labeling task~\cite{chen2015long,zhang2018neural}.
The core difference between these methods mainly lies in how they learn the contextual feature representation for each character in sentence.
For example, Zheng et al.~\cite{zheng2013deep} proposed to use multi-layer perceptrons to learn feature representations of characters from a fixed window.
Chen et al.~\cite{chen2015long} used LSTM to capture global contextual information.
They also explicitly captured the local context by combining the embedding of current character with the embeddings of neighbouring characters as the input of LSTM.
In~\cite{peng2017multi}, LSTM is used to learn character representations and CRF is used to jointly decode the labels.
These methods rely on a large number of labeled sentences to train CWS models and cannot exploit the useful information in Chinese dictionaries~\cite{zhang2018neural}.
Since there are massive Chinese words which are scarce or absent in the labeled sentences, these neural CWS methods usually have difficulty in correctly segmenting sentences containing these words~\cite{zhang2018neural}.

Recently, incorporating the dictionary information into neural Chinese word segmentation has attracted increasing attentions~\cite{yang2017neural,zhang2018neural}.
For example, Yang et al.~\cite{yang2017neural} proposed to incorporate external information such as punctuation, automatic segmentation and POS data into neural CWS via pretraining.
However, the useful information in Chinese dictionaries is not considered in their method.
Zhang et al.~\cite{zhang2018neural} proposed to incorporate the dictionary information into an LSTM based neural CWS method via feature engineering.
They used several handcrafted templates to build an additional feature vector for each character using the dictionary and the neighbouring characters.
These additional feature vectors are fed to another LSTM network to learn additional character representations.
However, designing these handcrafted feature templates needs a lot of domain knowledge.
In addition, more model parameters are introduced in their method, making it more difficult to train neural CWS model especially when training data is insufficient.
Different from~\cite{zhang2018neural}, our method to incorporate dictionary information into neural CWS can be trained in an end-to-end manner and does not need manual feature engineering.
Experimental results show that our approach can achieve better performance than the method in~\cite{zhang2018neural}.

\section{Our Approach}

In this section we first present the basic neural architecture for Chinese word segmentation used in our approach.
Then, we introduce our methods of incorporating dictionary information for neural CWS.

\subsection{Basic Neural Architecture}

Following many previous works~\cite{chen2015long,zhang2018neural}, in this paper we model Chinese word segmentation as a character-level sequence labeling problem.
For each character in a sentence, our model will assign one of the tags in a predefined tag set to it, indicating its position in a word.
We use the \emph{BMES} tagging scheme, where \emph{B}, \emph{M} and \emph{E} mean the beginning, middle and end position in the word, and \emph{S} represents single character word.

The basic neural architecture for CWS used in our approach is CNN-CRF.
This neural architecture contains three main layers.
The first layer is the character embedding layer.
In this layer, the input sentence is converted to a sequence of vectors.
Denote the input sentence as $\mathbf{x}=[c_1,c_2,...,c_M]$, where $M$ is the sentence length and $c_i$ is the $i$-th character in this sentence.
After the embedding layer, the input sentence will become $\mathbf{x}=[\mathbf{c}_1,\mathbf{c}_2,...,\mathbf{c}_M]$, where $\mathbf{c}_i \in \mathcal{R}^{D}$ is the embedding of character $c_i$ and $D$ is the embedding dimension.

The second layer is the CNN layer.
Previous studies show that local context information is important for Chinese word segmentation~\cite{cai2017fast,yang2017neural}.
In addition, many researchers have shown that CNN is effective in capturing local context information~\cite{lecun2015deep,dos2014deep,zhang2015character}.
Motivated by these observations, we use CNN to learn the contextual representations of characters for CWS.
Denote $\mathbf{w} \in \mathcal{R}^{KD}$ as the parameter of a filter with kernel size $K$, then the hidden representation of the $i$-th character generated by this filter is formulated as follows:
\begin{equation}
h_i=f(\mathbf{w}^T\times \mathbf{c}_{i-\lceil \frac{k-1}{2}\rceil:i+\lfloor \frac{k-1}{2}\rfloor}+b),
\end{equation}
where $\mathbf{c}_{i-\lceil \frac{k-1}{2}\rceil:i+\lfloor \frac{k-1}{2}\rfloor}$ is the concatenation of the embeddings of neighbouring characters, $f$ is the ReLU function, and $\mathbf{w}$ and $b$ are the parameters of the filter.
Multiple filters with different kernel sizes are used.
The final hidden representation of the $i$-th character is the concatenation of the output of all filters at this position, which is denoted as $\mathbf{h}_i \in \mathcal{R}^{F}$ ($F$ is the number of filters).

The third layer is the CRF layer.
In Chinese word segmentation there are usually strong dependencies among neighbouring tags~\cite{chen2015long}.
For example, the tag \emph{M} cannot follow tag \emph{S} or \emph{E}.
Following many previous works on CWS~\cite{peng2017multi,zhang2018neural}, we use CRF to capture the dependencies among neighbouring tags.
Denote the input sentence as $\mathbf{x}=[c_1,c_2,...,c_M]$, and the predicted tag sequence as $\mathbf{y}=[y_1,y_2,...,y_M]$, then the score of this prediction is formulated as:
\begin{equation}
g(\mathbf{x},\mathbf{y})=\sum_{i=1}^M (S_{i,y_i}+A_{y_{i-1},y_i}),
\end{equation}
where $S_{i,y_i}$ is the score of assigning tag $y_i$ to the $i$-th character, and $A_{y_{i-1},y_i}$ is the score of jumping from tag $y_{i-1}$ to tag $y_i$.
In our approach, $S_{i}$ is defined as:
\begin{equation}
S_{i} = \mathbf{W}^T\mathbf{h}_i+\mathbf{b},
\end{equation}
where $\mathbf{h}_i$ is the hidden representation of the $i$-th character learned by the CNN layer, and $\mathbf{W} \in \mathcal{R}^{F\times T}$ and $\mathbf{b} \in \mathcal{R}^{T}$ ($T$ is the size of the tag set) are the parameters for character score prediction.
In CRF, the probability of sentence $\mathbf{x}$ having tag sequence $\mathbf{y}$ is defined as:
\begin{equation}\label{eq.posterior}
\begin{split}
  p(\mathbf{y}|\mathbf{x}) = \frac{ \exp(g(\mathbf{x},\mathbf{y}))}{ \sum_{\mathbf{y}^\prime \in \mathcal{Y}(\mathbf{x})}\exp(g(\mathbf{x},\mathbf{y}^\prime))},
\end{split}
\end{equation}
where $\mathcal{Y}(\mathbf{x})$ is the set of all possible tag sequences of sentence $\mathbf{x}$.

Then the loss function can be formulated as:
\begin{equation}\label{eq.loss}
\begin{split}
\mathcal{L} = - \sum_{i=1}^{N} \log(p(\mathbf{y}_i|\mathbf{x}_i)),
\end{split}
\end{equation}
where $N$ is the number of labeled sentences for training, and $\mathbf{y}_i$ is the ground-truth tag sequence of the $i$-th sentence.

For prediction, given a sentence $\mathbf{x}$ to be segmented, the predicted tag sequence $\mathbf{y}^\star$ is the one with the highest likelihood:
\begin{equation}\label{eq.prediction}
\begin{split}
\mathbf{y}^\star = \mathop{\arg\max}_{\mathbf{y} \in \mathcal{Y}(\mathbf{x})} p(\mathbf{y}|\mathbf{x}).
\end{split}
\end{equation}
We use Viterbi algorithm to solve the decoding problem in Eq.~(\ref{eq.prediction}).

\subsection{Incorporating Dictionary Information for Neural CWS}

Existing neural CWS methods usually rely on a large number of labeled sentences for model training.
Researchers have found that the neural models trained on labeled sentences usually have difficulties in segmenting sentences which contain OOV or rarely appearing words~\cite{zhang2018neural}.
For example, a Chinese sentence is ``人工智能最近很火" (Recently AI is hot).
Its ground-truth segmentation is ``人工智能/最近/很火".
However, if ``人工智能" (AI) does not appear in the labeled data or only appears for a few times, then there is a large probability that this sentence will be segmented into ``人工/智能/最近/很火", since ``人工" and ``智能" are both popular words which may frequently appear in the labeled data.
Luckily, many of these rare words are included in Chinese dictionary.
If the neural model is aware of that ``人工智能" is a Chinese word, then it can better segment the aforementioned sentence.
Thus, dictionary information has the potential to improve the performance of neural CWS methods.

In this paper we propose two methods for incorporating dictionary information into training neural CWS models.
Next we will introduce them in detail.

\subsubsection{Pseudo Labeled Data Generation}

Our first method for incorporating dictionary information into neural CWS model training is based on pseudo labeled data generation.
More specifically, given a Chinese dictionary which contains a list of Chinese words, we randomly sample $U$ words and use them to form a pseudo sentence.
For example, assuming that three words ``很火", ``最近" and ``人工智能" are sampled, then a pseudo sentence ``很火最近人工智能" can be built.
Since the boundaries of these words are already known, the tag sequence of the generated pseudo sentence can be automatically inferred.
For instance, the tag sequence of aforementioned pseudo sentence is ``B/E/B/E/B/M/M/E" under the \emph{BMES} tagging scheme.
Then we repeat this process until $N_p$ pseudo labeled sentences are generated.
These pseudo labeled sentences are added to labeled data set to enhance the training of neural CWS model.

Since the pseudo labeled sentences may have different informativeness from the manually labeled sentences, we assign different weights to the loss on these two kinds of training data, and the final loss function is formulated as:
\begin{equation}\label{eq.loss_pseudo}
\begin{split}
\mathcal{L} = - \sum_{i=1}^{N} \log(p(\mathbf{y}_i|\mathbf{x}_i))- \lambda_1 \sum_{i=1}^{N_p} \log(p(\mathbf{y}^s_i|\mathbf{x}^s_i)),
\end{split}
\end{equation}
where $\mathbf{x}^s_i$ and $\mathbf{y}^s_i$ represent the $i$-th pseudo labeled sentence and its tag sequence, and $\lambda_1$ is a non-negative coefficient.

\subsubsection{Multi-task Learning}

\begin{figure}
  \centering
  \includegraphics[width=0.35\textwidth]{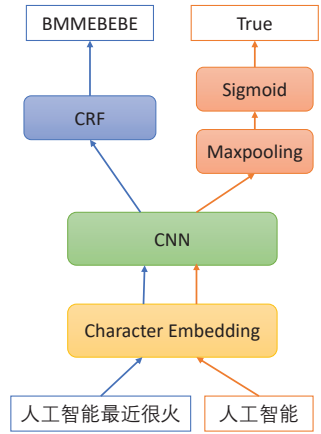}
  \caption{Our proposed framework for jointly training CWS and word classification models. The left part is for CWS and the right part is for word classification.}
  \label{joint}
\end{figure}

Our second method for incorporating dictionary information into neural CWS model training is based on multi-task learning.
In this method, we design an additional task, i.e., word classification, which means classifying a sequence of Chinese characters based on whether it can be a Chinese word.
For example, the character sequence ``人工智能" will be classified to be true, while the character sequence ``人重智新" will be classified to be false.
The positive samples are obtained from a Chinese dictionary.
The negative samples are obtained via randomly sampling a word from the dictionary, and then each character in this word will be randomly replaced by a random selected character with a probability $p$.
This step is repeated multiple times until a predefined number of negative samples are obtained.
We use a neural method for the word classification task, whose architecture is similar with the CNN-CRF architecture for CWS, except that the CRF layer is replaced by a max-pooling layer and a sigmoid layer for binary classification.
The loss function of the word classification task is formulated as:
\begin{equation}\label{eq.word}
\begin{split}
\mathcal{L} = \sum_{i=1}^{N_w} \log(1+e^{-y_i s_i}),
\end{split}
\end{equation}
where $N_w$ is the number of training samples for word classification, $s_i$ is the predicted score of the $i$-th sample, and $y_i$ is the word classification label which can be 1 or -1 (1 represents true and -1 represents false).

Motivated by multi-task learning, we propose a unified framework to jointly train the Chinese word segmentation model and the word classification model, which is illustrated in Fig.~\ref{joint}.
In our framework, the CWS model and the word classification model share the same embedding layer and CNN layer.
In this way, these two layers can better capture the word information in Chinese dictionary via jointly training with the word classification task, and the performance of CWS can be improved.
In model training we assign different weights to the loss of these two tasks, and the final loss function is:
\begin{equation}\label{eq.loss_multi}
\begin{split}
\mathcal{L} = -(1-\lambda_2) \sum_{i=1}^{N} \log(p(\mathbf{y}_i|\mathbf{x}_i))+ \lambda_2 \sum_{i=1}^{N_w} \log(1+e^{-y_i s_i}),
\end{split}
\end{equation}
where $\lambda_2$ is a coefficient ranging from 0 to 1.

\section{Experiment}
\subsection{Dataset}
In our experiments we used two benchmark datasets released by the third international Chinese language processing bakeoff\footnote{\url{http://sighan.cs.uchicago.edu/bakeoff2006/download.html}}~\cite{levow2006third}.
The detailed statistics of these two datasets are summarized in Table~\ref{dataset}.
We used the last 10\% data of the training set as development set.
\begin{table*}[htbp]
  \caption{The statistics of datasets.}
  \label{dataset}
  \centering
  \begin{tabular}{@{\extracolsep{0pt}} c|c|c|c|c|c}
    \Xhline{1pt}
    \multicolumn{2}{c|}{Dataset} & \#Sentence & \#Word & \#Character & OOV Rate\\
    \hline
    \multirow{2}{*}{MSRA} & Train & 46.3K & 1.27M & 2.17M & - \\
    \cline{2-6}
    & Test & 4.4K & 0.10M & 0.17M & 3.4\% \\
    \hline
    \multirow{2}{*}{UPUC} & Train & 18.8K & 0.51M & 0.83M & - \\
    \cline{2-6}
    & Test & 5.1K & 0.15M & 0.26M & 8.8\% \\
  \Xhline{1pt}
\end{tabular}
\end{table*}
\subsection{Experimental Settings}
The character embeddings used in our experiments were pretrained on the Sogou news corpus\footnote{\url{http://www.sogou.com/labs/resource/ca.php}} using the word2vec\footnote{\url{https://code.google.com/archive/p/word2vec/}} tool.
The dimension of character embedding is 200.
We used 400 filters in the CNN layer and the kernel sizes of these filters range from 2 to 5.
Rmsprop~\cite{dauphin2015equilibrated} was used as the algorithm for neural model training.
The learning rate was set to 0.001 and the batch size was 64.
Dropout was applied to the embedding layer and the CNN layer.
The dropout rate was set to 0.3.
We use early stopping strategy.
When the loss on the development set doesn't reduce after 3 consecutive epochs, the training is stopped.
We repeated each experiment for 5 times and reported the average results.

\subsection{Performance Evaluation}

In this section we compare our approach with several baseline methods.
These baseline methods include:
(1) Chen et al. \cite{chen2015long}, a LSTM based CWS method which also considers local contexts;
(2) LSTM-CRF, a popular neural CWS method based on the LSTM-CRF architecture~\cite{peng2017multi,zhang2018neural};
(3) CNN-CRF, a neural CWS method based on the CNN-CRF architecture, which is the basic model for our approach;
(4) Zhang et al. \cite{zhang2018neural}, a neural CWS method which can incorporate dictionary information via feature templates.
In order to evaluate the performance of different methods under different amounts of labeled data, we randomly sampled different ratios of labeled data for training.
The experimental results are summarized in Tables~\ref{table.MSR} and~\ref{table.UPUC}.
\begin{table*}[!htb]
  \caption{The performance of different methods on the \emph{MSRA} dataset. $P$, $R$ and $F$ represent precision, recall and Fscore respectively. \emph{Ours\_Pseudo} represents our approach based on pseudo labeled data generation, and \emph{Ours\_Multi} represents our approach based on multi-task learning.}\label{table.MSR}
  \centering
  \resizebox{0.9\textwidth}{!}{
    \begin{tabular}{@{\extracolsep{0pt}} c| c| c| c| c| c| c| c| c| c}
      \Xhline{1pt}
      \multirow{2}{*}{} &
      \multicolumn{3}{c|}{1\%} & \multicolumn{3}{c|}{10\%} & \multicolumn{3}{c}{100\%} \\
      \Xcline{2-10}{0.5pt}
      & $P$ & $R$  & $F$ & $P$ & $R$  & $F$ & $P$ & $R$  & $F$ \\
      \Xhline{0.9pt}
        Chen et al. \cite{chen2015long}  & 75.50 & 75.80 & 75.64 & 87.71 & 86.22 & 86.96 & 94.24 & 93.35 & 93.80 \\
      \hline
      LSTM-CRF  & 75.88 & 74.86 & 75.36 & 85.52 & 84.81 & 85.16 & 94.26 & 93.29 & 93.78 \\
      \hline
      CNN-CRF  & 75.59 & 74.43 & 75.00 & 89.72 & 89.14 & 89.43 & 95.03 & 94.53 & 94.78 \\
      \hline
      \hline
      Zhang et al. \cite{zhang2018neural}  & 75.75 & 75.95 & 75.85 & 89.52 & 89.01 & 89.27 & 95.71 & 95.41 & 95.56 \\
      \hline
      \hline
      Ours\_Pseudo  & 80.58 & 77.97 & 79.25 & 90.49 & 89.59 & 90.04 & 95.36 & 94.71 & 95.03 \\
      \hline
      Ours\_Multi  & 78.47 & 77.31 & 77.88 & 89.91 & 89.27 & 89.59 & 95.10 & 94.50 & 94.80\\
      \Xhline{1pt}
    \end{tabular}
  }
\end{table*}
\begin{table*}[!htb]
  \caption{The performance of different methods on the \emph{UPUC} dataset.}\label{table.UPUC}
  \centering
  \resizebox{0.9\textwidth}{!}{
    \begin{tabular}{@{\extracolsep{0pt}} c| c| c| c| c| c| c| c| c| c}
      \Xhline{1pt}
      \multirow{2}{*}{} &
      \multicolumn{3}{c|}{5\%} & \multicolumn{3}{c|}{25\%} & \multicolumn{3}{c}{100\%} \\
      \Xcline{2-10}{0.5pt}
      & $P$ & $R$  & $F$ & $P$ & $R$  & $F$ & $P$ & $R$  & $F$  \\
      \Xhline{0.9pt}
        Chen et al. \cite{chen2015long}  & 82.31 & 82.60 & 82.44 & 88.00 & 89.90 & 88.94 & 90.79 & 92.92 & 91.84 \\
      \hline
      LSTM-CRF  & 81.08 & 80.88 & 80.98 & 86.76 & 88.40 & 87.57 & 91.39 & 92.58 & 91.98 \\
      \hline
      CNN-CRF  & 82.44 & 84.50 & 83.46 & 89.95 & 91.57 & 90.75 & 92.22 & 93.84 & 93.02 \\
      \hline
      \hline
      Zhang et al. \cite{zhang2018neural}  & 83.38 & 84.98 & 84.17 & 89.93 & 91.41 & 90.66 & 92.60 & 93.89 & 93.24 \\
      \hline
      \hline
      Ours\_Pseudo  & 87.37 & 86.56 & 86.97 & 90.97 & 92.04 & 91.50 & 92.77 & 94.09 & 93.43 \\
      \hline
      Ours\_Multi  & 84.59 & 86.22 & 85.40 & 90.43 & 91.68 & 91.05 & 92.35 & 93.93 & 93.13\\
      \Xhline{1pt}
    \end{tabular}
  }
\end{table*}
According to Tables~\ref{table.MSR} and~\ref{table.UPUC}, we have two observations.

First, both of our approaches perform better than various neural CWS methods which do not consider dictionary information, and the performance advantage becomes larger when training data is insufficient.
This result validates that by incorporating the dictionary information our approaches can effectively improve the performance of neural CWS.
This is because there are many words which do not appear or rarely appear in the training data, and the neural CWS models which are trained purely on labeled data usually have difficulty in segmenting sentences containing these words.
Many of these words are usually included in Chinese dictionaries, and exploiting the useful information in dictionaries can help the neural CWS model better recognize these words.

Second, although the method proposed in \cite{zhang2018neural} can also incorporate the dictionary information for CWS, our approaches usually can outperform it, especially when training data is insufficient.
This result shows that our approaches are more appropriate for incorporating dictionary information for CWS than the method proposed in \cite{zhang2018neural}.
This is maybe because in \cite{zhang2018neural} the feature templates for incorporating dictionary information are manually designed, which may not be optimal.
In addition, in \cite{zhang2018neural} an additional LSTM network is used to learn character representations from these dictionary based features.
Thus, more model parameters are incorporated, making it more difficult to train the CWS model especially when training data is insufficient.
Our approaches do not rely on feature engineering and the additional model parameters introduced in our approaches are limited.
Thus, our approach can achieve better performance than \cite{zhang2018neural}.

\subsection{Influence of Dictionary}

In this section we conducted several experiments to explore the influence of the type and the size of Chinese dictionary on the performance of our approach.
\begin{figure}[htbp]
\begin{minipage}[t]{0.48\textwidth}
  \centering
  \includegraphics[width=6cm]{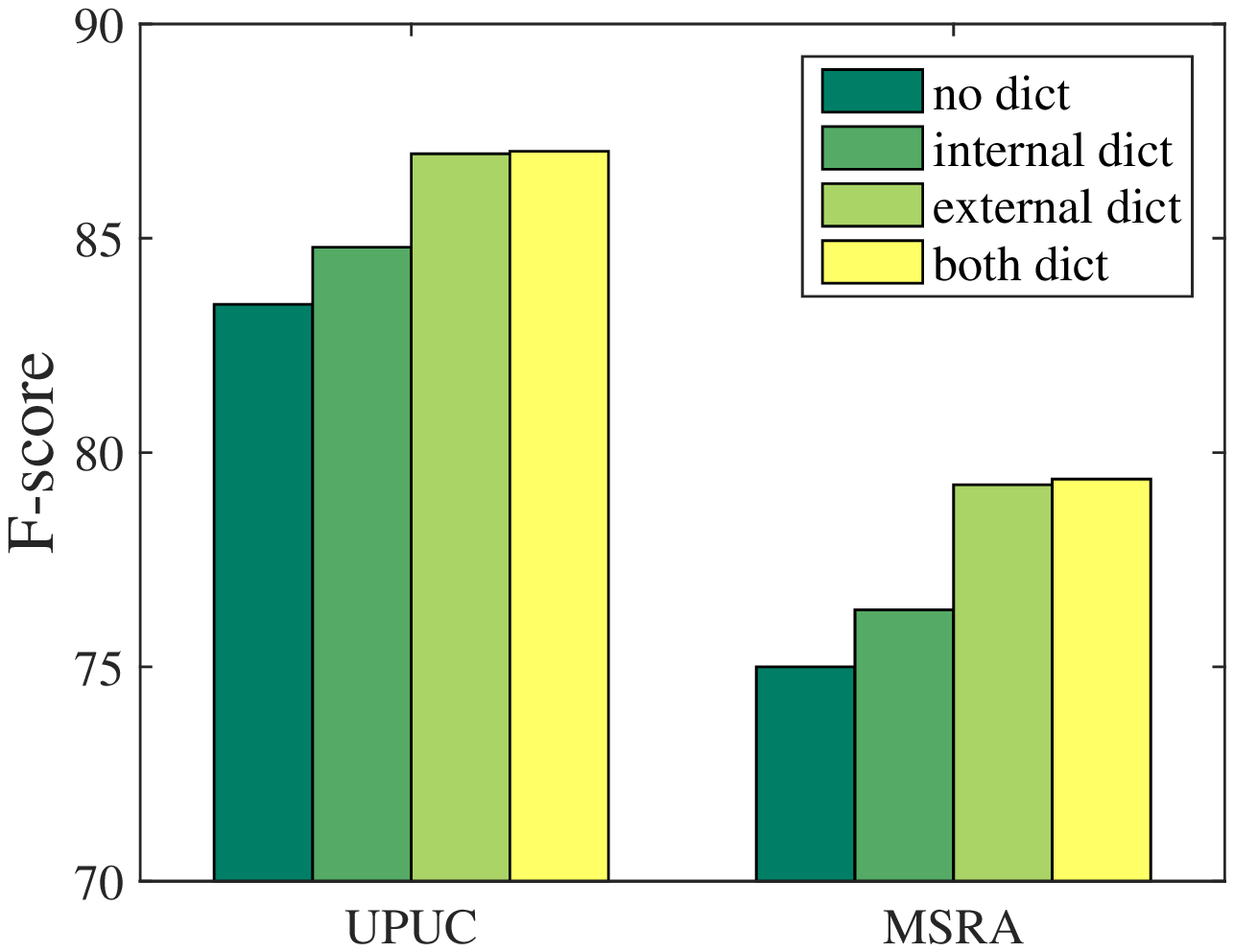}
  \caption{The influence of dictionary type.}
  \label{idtype}
\end{minipage}
\begin{minipage}[t]{0.48\textwidth}
  \centering
  \includegraphics[width=6cm]{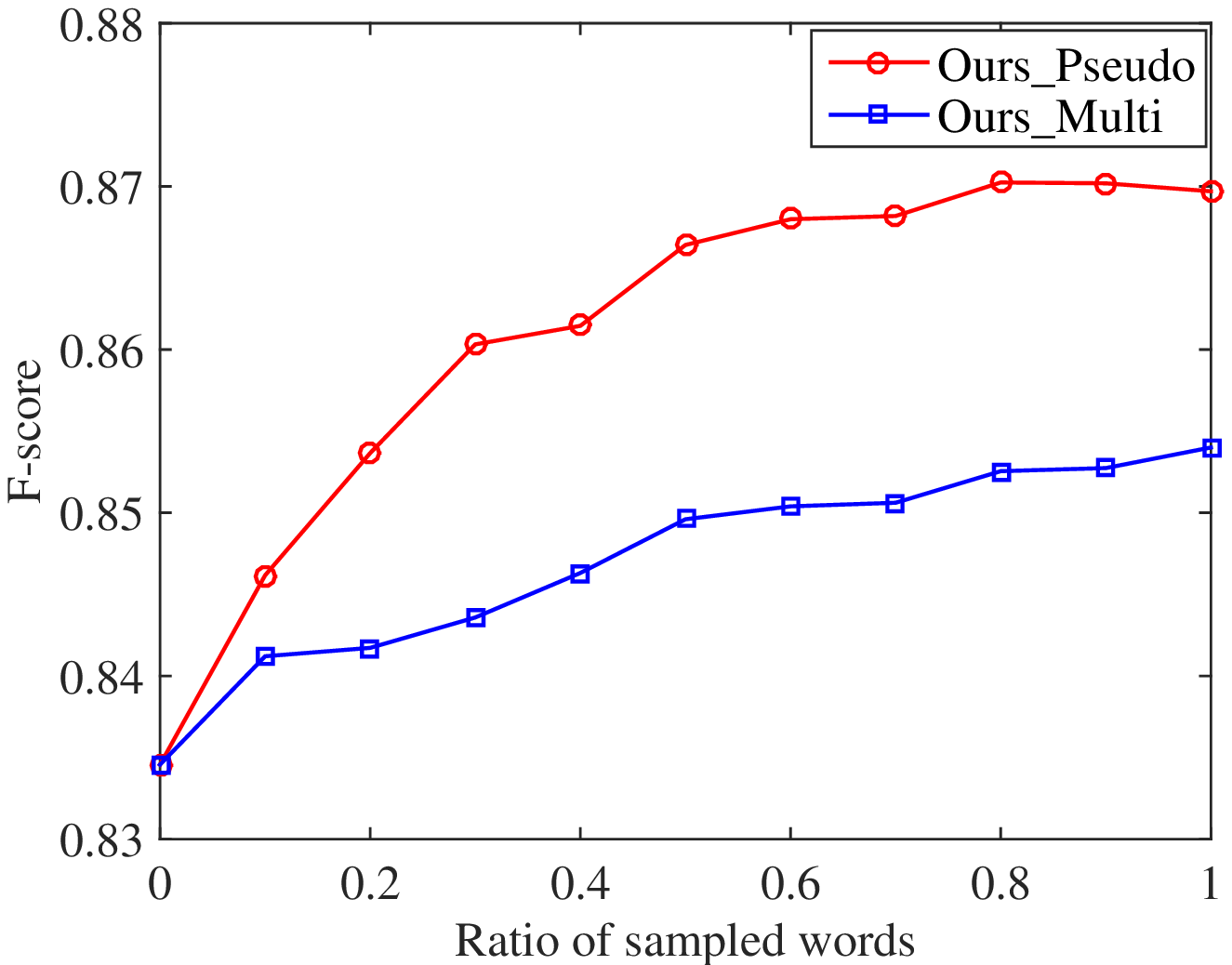}
  \caption{The influence of dictionary size.}
  \label{idsize}
\end{minipage}
\end{figure}

First, we explore the influence of dictionary type.
In previous section, the Chinese dictionary used in our approach is the Sogou Chinese Dictionary, which can be regarded as an external dictionary.
We also built an internal dictionary using the words appearing in the training data.
The results of our approach without any dictionary, with only internal dictionary, with only external dictionary, and with both dictionaries are summarized in Fig.~\ref{idtype}.
We randomly sampled 5\% training data of \emph{UPUC} dataset and 1\% training data of \emph{MSRA} dataset for model training.

According to Fig.~\ref{idtype}, with external dictionary our approach can improve the performance of CWS.
In addition, our approach can also improve the performance with only internal dictionary.
This result is promising, since the internal dictionary is built on the words appearing in training data and no external resource is involved.
In addition, incorporating both internal and external dictionaries can further improve the performance of our approach, which indicates that these two dictionaries contain complementary information.

Next, we explore the influence of the dictionary size on the performance of our approach.
We randomly sampled different numbers of words from the Sogou dictionary, and the experimental results on \emph{UPUC} dataset are summarized in Fig.~\ref{idsize}.

From Fig.~\ref{idsize}, with the size of dictionary grows the performance improves.
This result is intuitive since when a dictionary contains more words it can have a better coverage of the Chinese words, and our approach can benefit from this by incorporating the useful information in these words into training neural CWS model.
\subsection{The Influence of Parameters}
\begin{figure}[htbp]
\begin{minipage}[t]{0.48\textwidth}
  \centering
  \includegraphics[width=6cm]{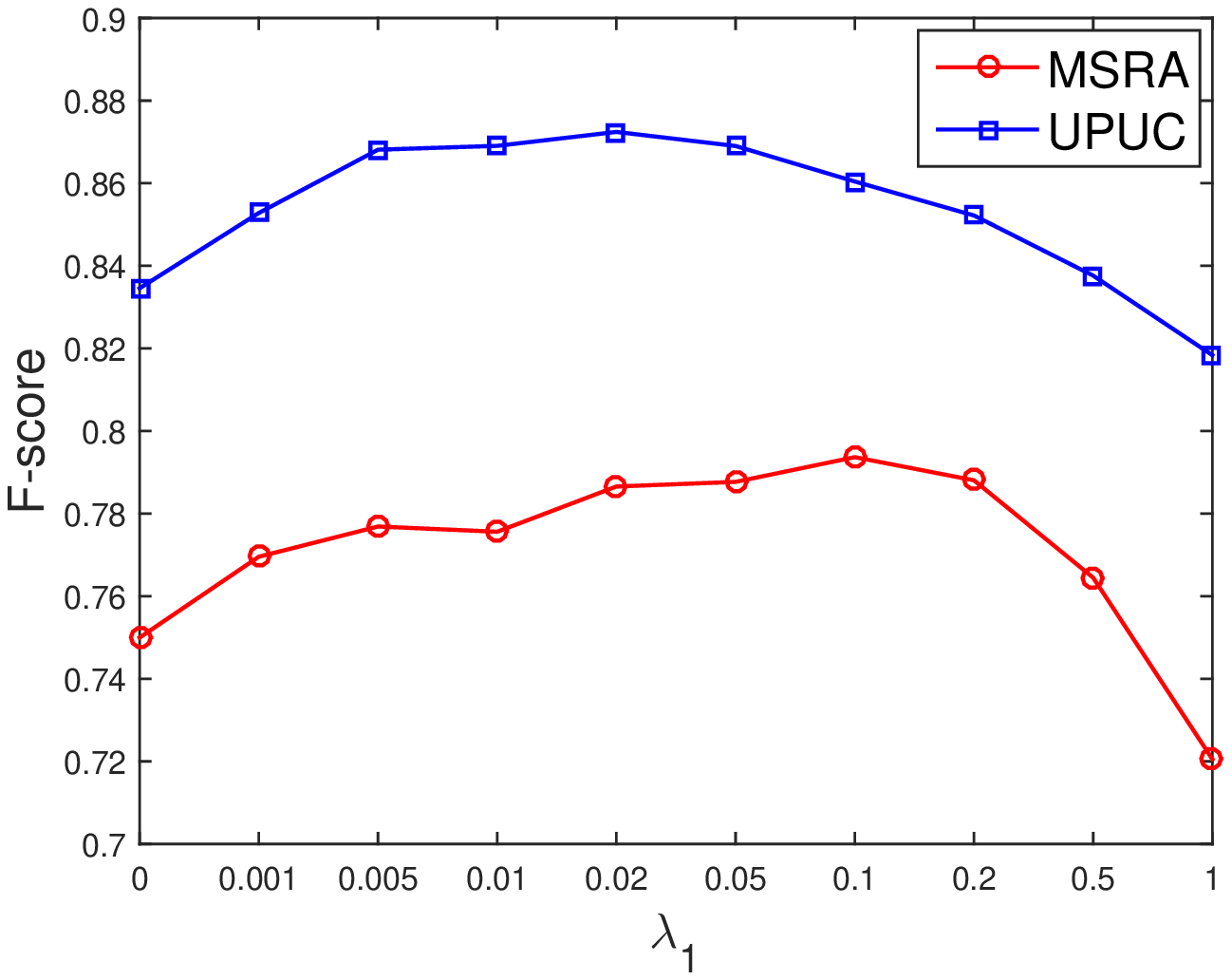}
  \caption{The influence of $\lambda_1$.}
  \label{ilambda1}
\end{minipage}
\begin{minipage}[t]{0.48\textwidth}
  \centering
  \includegraphics[width=6cm]{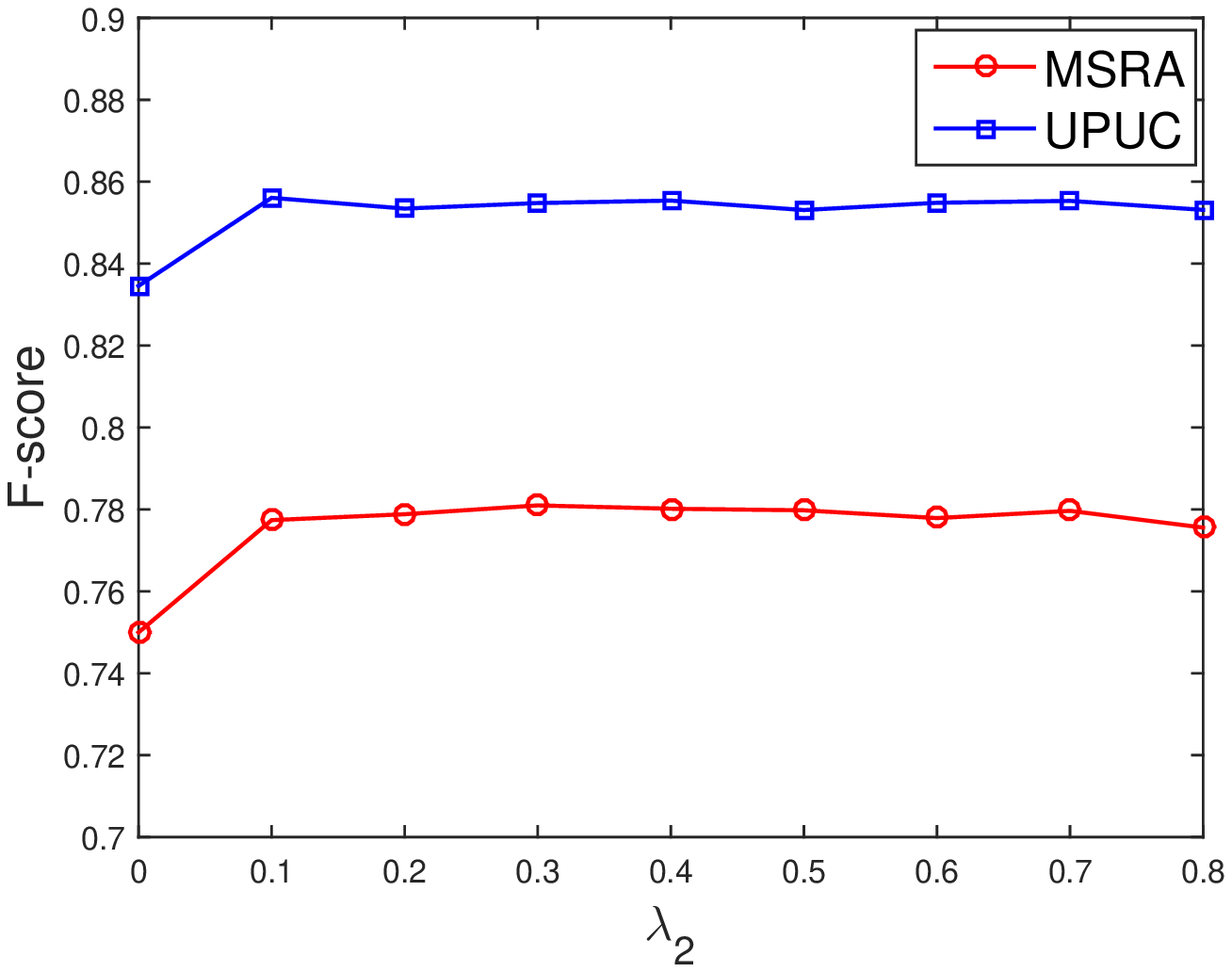}
  \caption{The influence of $\lambda_2$.}
  \label{ilambda2}
\end{minipage}
\end{figure}
There are two most important parameters in our approaches.
The first one is $\lambda_1$, which controls the relative importance of pseudo labeled samples.
The second one is $\lambda_2$, which controls the relative importance of word segmentation task.
The influence of these parameters on the performance of our approaches is illustrated in Fig.~\ref{ilambda1} and Fig.~\ref{ilambda2}.

From Fig.~\ref{ilambda1} and Fig.~\ref{ilambda2} we can see that when $\lambda_1$ and $\lambda_2$ are too small, the performance of our approach is not optimal, and improves as $\lambda_1$ and $\lambda_2$ increase.
This is because when these parameters are too small, the useful information in the dictionary is not fully exploited.
However, when $\lambda_1$ and $\lambda_2$ become too large, the performance of our approach decreases.
This is because in these cases the pseudo labeled samples and the word classification task are over-emphasized.
Accordingly, the manually labeled samples and the CWS task are not fully respected.
Thus, a moderate value is most appropriate for $\lambda_1$ and $\lambda_2$.

\subsection{Case Study}
\begin{table*}[htbp]
  \caption{Several Chinese word segmentation examples.}
  \label{case}
  \centering
  \begin{tabular}{@{\extracolsep{0pt}} c|c|c}
    \Xhline{1pt}
     & Example 1 & Example 2 \\
    \hline
    Original & ５名男子和被害人有恩怨 & 警方一口气带回了５０多人 \\
    \hline
    CNN-CRF & ５/名/男子/和/被/害/人/有/恩怨 & 警方/一/口/气/带回/了/５０多/人 \\
    \hline
    +Internal dictionary & ５/名/男子/和/被/害/人/有/恩怨 & 警方/一口气/带回/了/５０多/人 \\
    \hline
    +External dictionary & ５/名/男子/和/被害人/有/恩怨 & 警方/一口气/带回/了/５０多/人 \\
  \Xhline{1pt}
\end{tabular}
\end{table*}

In this section we conducted several case studies to explore why our approach can improve the performance of Chinese word segmentation via incorporating the dictionary information.
Several segmentation results of our approach without dictionary (i.e., the CNN-CRF method), with internal dictionary and with external dictionary are shown in Table~\ref{case}.
For illustration purpose, we only show the results of our approach based on pseudo labeled data generation.

According to Table~\ref{case}, after incorporating the dictionary information, our approach can correctly segment many sentences where the basic CNN-CRF method has difficulties.
For instance, in the first example, the true segmentation of ``被害人" is ``被害人".
However, CNN-CRF incorrectly segments it into ``被/害/人", because ``被害人" is an OOV word which does not appear in training data.
Our approach with external dictionary can correctly segment this sentence because the word ``被害人" is in the external dictionary and our approach can fully exploit this useful information.
In the second example, CNN-CRF incorrectly segments ``一口气" into ``一/口/气", because ``一口气" is an rare word which only appears 2 times in the training data which is difficult for neural CWS model to segment it.
Since this word is in both internal and external dictionaries, our approach with either dictionary can correctly segment this sentence.
Thus, these results clearly show that incorporating dictionary information into training neural CWS methods is beneficial.

\section{Conclusion}
In this paper we present two approaches for incorporating the dictionary information into neural Chinese word segmentation.
The first one is based on pseudo labeled data generation, where pseudo labeled sentences are generated by combining words randomly sampled from dictionary.
The second one is based on multi-task learning, where we design a word classification task and using the dictionary to build labeled samples.
We jointly train the Chinese word segmentation and the word classification task via sharing the same network parameters.
Experimental results on two benchmark datasets show that our approach can effectively improve the performance of Chinese word segmentation, especially when training data is insufficient.

\bibliographystyle{splncs04}
\bibliography{mybibliography}





\end{CJK*}
\end{document}